\documentclass[sigconf]{acmart}

\usepackage{booktabs} 
\usepackage{comment}
\usepackage{amsmath}
\usepackage{graphicx}
\usepackage{booktabs}
\usepackage{multirow}
\usepackage{adjustbox}
\setcopyright{rightsretained}

\begin{document}
\title{ViLP:  Knowledge Exploration using Vision, Language, and Pose Embeddings for Video Action Recognition}

\author{Soumyabrata Chaudhuri}
\affiliation{
  \institution{Indian Institute of Technology Bhubaneswar}
  \city{Bhubaneswar}
  \state{Odisha}
  \country{India} \\
  \texttt{21CS01032@iitbbs.ac.in}
}
\author{Saumik Bhattacharya}
\affiliation{
  \institution{Indian Institute of Technology Kharagpur}
  \city{Kharagpur}
  \state{West Bengal}
  \country{India} \\
  \texttt{saumik@ece.iitkgp.ac.in}
}




\begin{abstract}
Video Action Recognition (VAR) is a challenging task due to its inherent complexities. Though different approaches have been explored in the literature, designing a unified framework to recognize a large number of human actions is still a challenging problem. Recently,  Multi-Modal Learning (MML) has demonstrated promising results in this domain. In literature, 2D skeleton or pose modality has often been used for this task, either independently or in conjunction with the visual information (RGB modality) present in videos. However, the combination of pose, visual information, and text attributes has not been explored yet, though text and pose attributes independently have been proven to be effective in numerous computer vision tasks. In this paper, we present the first pose augmented Vision-language model (VLM) for VAR. Notably, our scheme achieves an accuracy of 92.81\% and 73.02\% on two popular human video action recognition benchmark datasets, UCF-101 and HMDB-51, respectively, even without any video data pre-training, and an accuracy of 96.11\% and 75.75\% after kinetics pre-training.
\end{abstract}

%
%


\maketitle

\begin{figure}
    \centering
    \includegraphics[width=0.5\textwidth]{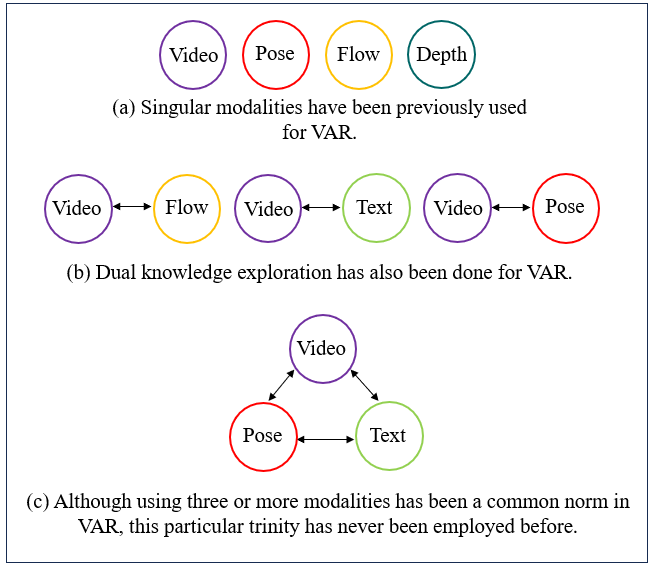}
    \caption{The key idea: In this work, we have simultaneously used video encoding, text encoding and pose encoding for VAR. }
    \label{fig:idea}
\end{figure}
\section{Introduction}
Human action recognition from video input is a well-known computer vision problem. Several different approaches using classical techniques, like temporal template matching \cite{bobick2001recognition}, optical flow-based motion model \cite{efros2003recognizing}, motion history volume \cite{weinland2006free}, as well as deep techniques, like 3D CNN \cite{ji20123d}, CNN-LSTM network \cite{baccouche2011sequential}, spatio-temporal skeleton model \cite{song2017end}. Previously, while designing VAR algorithms,  researchers had focused primarily on raw RGB video input to capture motion cues \cite{bobick2001recognition,efros2003recognizing,weinland2006free}, but soon additional modalities, like pose, skeleton, depth, etc. \cite{yao2011does, song2017end, chen2016real}, to improve the recognition performance. In recent years, the remarkable achievements of Multi-Modal Learning (MML) \cite{7346486,Gallo:2019:IVCNZ} in Computer Vision (CV) along with the emergence of large scale pre-training in Natural Language Processing (NLP) \cite{devlin2018bert,brown2020language,radford2019language,zhang2019ernie,raffel2020exploring} has led to the development of Vision-Language Models (VLMs) such as CLIP \cite{radford2021learning}, ALIGN \cite{jia2021scaling}, Florence \cite{yuan2021florence} and CoCa \cite{yu2022coca}. These models leverage the relationship between large scale image-text pairs and exhibit impressive adaptability in various vision tasks. 

Therefore, it is quite natural to take advantage of the transferability of these models during video encoding or text encoding for downstream tasks like human action recognition. In this paper, we focus on a multimodal approach that takes text encoding and pose encoding along with video data to boost the performance of the VAR task. As explored in some of the recent tasks, text encoding as  auxiliary information can greatly improve the overall performance of a deep model \cite{yuan2021florence,roy2022tips, deng2023llms}. Furthermore, our decision to use pose instead of other modalities like depth or optical flow is rooted in the belief that the location of key-points of the human body is an indispensable cue for ascertaining what type of action the people are performing \cite{lv2007single,zhang2019view}. As a result, we employ a combination of three encoders- video, text, and pose to represent each of the three modalities. In Fig. \ref{fig:idea}, we present the key idea of the proposed technique. As shown in Fig. \ref{fig:idea}, though several singular modalities and dual modalities have been explored for VAR, we have investigated the simultaneous contributions of video, pose, and text modalities for the same task.  

To demonstrate the efficacy of this particular amalgamation of modalites, we perform comprehensive experiments on two popular human action datasets, UCF-101 \cite{soomro2012ucf101} and HMDB-51 \cite{Kuehne2011HMDBAL}. As anticipated, the incorporation of the supplementary pose modality leads to enhanced performance across all scenarios. Interestingly, even in the absence of pre-training on any video dataset, such as kinetics, our approach achieves 92.81\% and 73.02\% accuracy on UCF-101 and HMDB-51, respectively.

Our main contributions can be summarized as follows:
\begin{itemize}
    \item To the best of our knowledge, this is the first attempt to integrate pose encoding and text encoding with video encoding for the VAR task. 
    \item The proposed pipeline achieves competitive results even without any pretraining on large databases like kinetics. This helps in much faster training, even in a resource-constrained setting. This proves the effectiveness of the auxiliary modalities for VAR. 
    \item Our proposed framework outperforms existing SOTA techniques with kinetics pretraining. 
\end{itemize}
The rest of the paper is organized as follows. In Sec. \ref{sec:related_works}, we discuss the existing methods for VAR and the importance of multiple modalities in the mentioned task. We discuss the proposed methodology in Sec. \ref{sec:method}. The experimental results for the proposed method, along with a detailed comparison with SOTA approaches, have been reported in Sec. \ref{sec:results}. Finally, we conclude the discussion in Sec. \ref{sec:conclusion} by discussing the key observations and the future scopes of the proposed work. 
\begin{figure*}
    \centering
    \includegraphics[width=\textwidth]{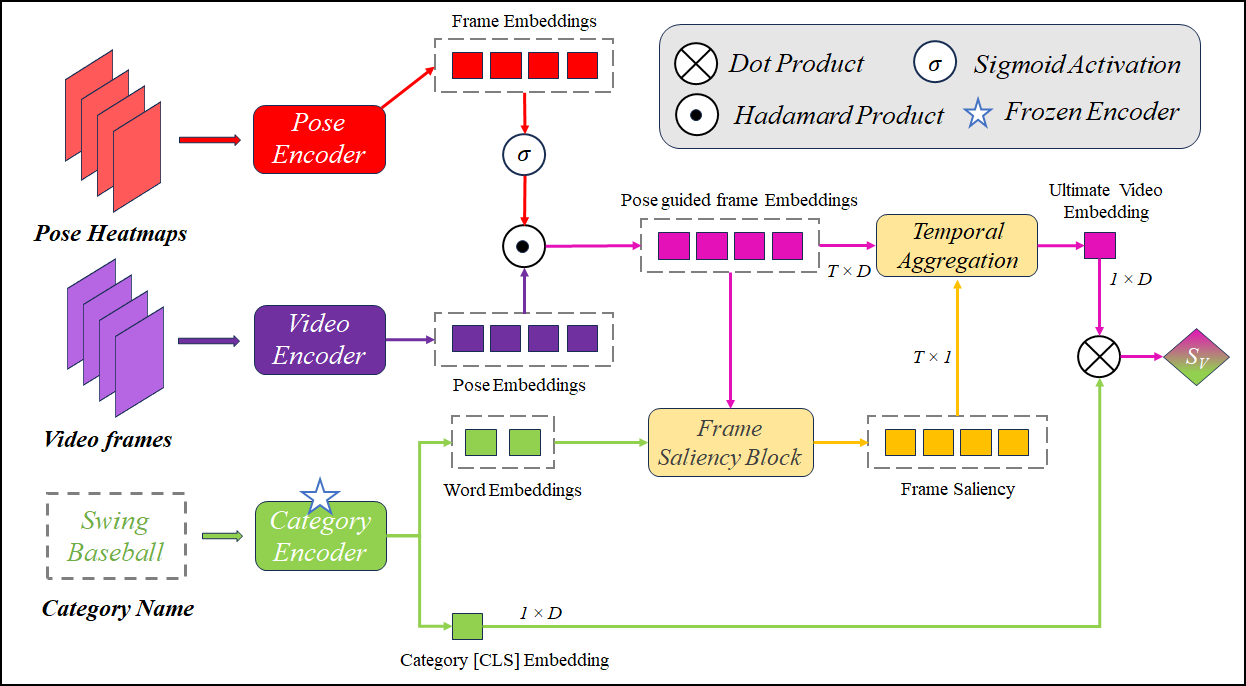}
    \caption{The proposed ViLP model: The proposed framework has three different input modalities, viz. pose branch, video branch and the category branch. The output of the model is the overall similarity score $S_V$. }
    \label{fig:model}
\end{figure*}
\section{Related Works} \label{sec:related_works}
{\bfseries Video Action Recognition:}
Throughout the last decade convolutional networks have been the go-to model for a variety of computer vision problems including VAR. Early works focused on learning spatial and temporal information in parallel \cite{Feichtenhofer_2019_ICCV,feichtenhofer2016spatiotemporal,simonyan2014two,wang2016temporal}. The approach in \cite{Feichtenhofer_2019_ICCV}, the SlowFast network, utilizes two pathways, Slow and Fast, to capture spatial semantics and motion at different frame rates, enabling efficient and accurate video recognition. \cite{feichtenhofer2016spatiotemporal} introduces spatiotemporal ResNets, a novel architecture that combines Two-stream Convolutional Networks (ConvNets) and Residual Networks (ResNets) for human action recognition in videos. The approach incorporates residual connections between appearance and motion pathways, allowing spatiotemporal interaction, and transforms pretrained image ConvNets into spatiotemporal networks with learnable convolutional filters for hierarchical learning of complex spatiotemporal features. \cite{wang2016temporal} presents TSN (Temporal Segment Network), a novel framework for video-based action recognition, leveraging long-range temporal structure modeling through sparse temporal sampling and video-level supervision, enabling efficient and effective learning with limited training samples. Subsequent works centred on improving temporal context modelling for 2D CNN backbones by developing plug-and-play temporal modules \cite{li2020tea,liu2020teinet,liu2021tam,qiu2017learning,tran2018closer}. Recently, with the introduction of Vision Transformer \cite{dosovitskiy2020image,han2021transformer,liu2021swin} for image recognition, a plethora of transformers are emerging in the video domain as well, such as VideoSwin \cite{Liu_2022_CVPR}, TimeSFormer \cite{gberta_2021_ICML}, ViViT \cite{arnab2021vivit} and MViT \cite{fan2021multiscale}.

{\bfseries Usage of Multiple Modalities:}
Two stream I3D network \cite{simonyan2014two} fused RGB modality with optical flow, trying to leverage both the spatial as well as the motion information present in frames. \cite{7346486} handled the complexity of actions by dividing it into the kinetics of body parts and analyzing the actions based on these partial descriptors. Their joint sparse regression-based learning method utilized the structured sparsity to model each action as a combination of multimodal features from a sparse set of body parts and employed a heterogeneous set of depth and skeleton-based features to represent the dynamics and appearance of parts. With the popularisation of Vision-language pre-training using large scale image-text pairs \cite{radford2021learning,fang2023uatvr,luo2022clip4clip,cap4video,zhao2022centerclip}, powerful models with the ability to align visual representation with rich linguistic semantics are arising. In recent years, this combination of vision and language has also been used for action recognition in videos \cite{bike,text4vis}, and they have yielded interesting results. However, to the best of our knowledge, there has not been an attempt to fruitfully utilize the rich triad of vision, language, and pose. Our approach is to augment video encoding with a textual encoding that provides auxiliary information about the present action and attribute with a third pose branch which will seek to model the skeletal information present in human action videos, resulting in an effective merging of information between the three modalities.

{\bfseries Human Pose Estimation:}
Human pose estimation(HPE) involves predicting the spatial locations of human body joints from images or videos. It plays a crucial role in various applications such as action recognition, human-computer interaction, and pose-based medical diagnosis. Over the years, significant progress has been made in this field, with several approaches achieving promising results.
One of the early influential works is the DeepPose model proposed by Toshev and Szegedy in 2014 \cite{toshev2014deeppose}. This work demonstrated the efficacy of deep learning in estimating human poses accurately.
Following the success of DeepPose, various CNN-based architectures have been proposed to further improve pose estimation accuracy. Newell et al. introduced the Stacked Hourglass model in 2016 \cite{newell2016stacked}, which employs a series of hourglass modules to capture multi-scale information and refine the pose predictions iteratively, leading to more accurate results.
In recent years, there has been a surge in attention towards one-stage pose estimation models. In 2018 \cite{Xiao_2018_ECCV} proposed a one-stage model that achieves competitive performance with significantly fewer parameters, making it more efficient for real-time applications.
In this paper we follow openpose algorithm \cite{cao2017realtime}, which utilizes part affinity fields to model relationships between body joints for handling complex poses involving multiple individuals.
\begin{figure}
    \centering
    \includegraphics[width=0.9\linewidth]{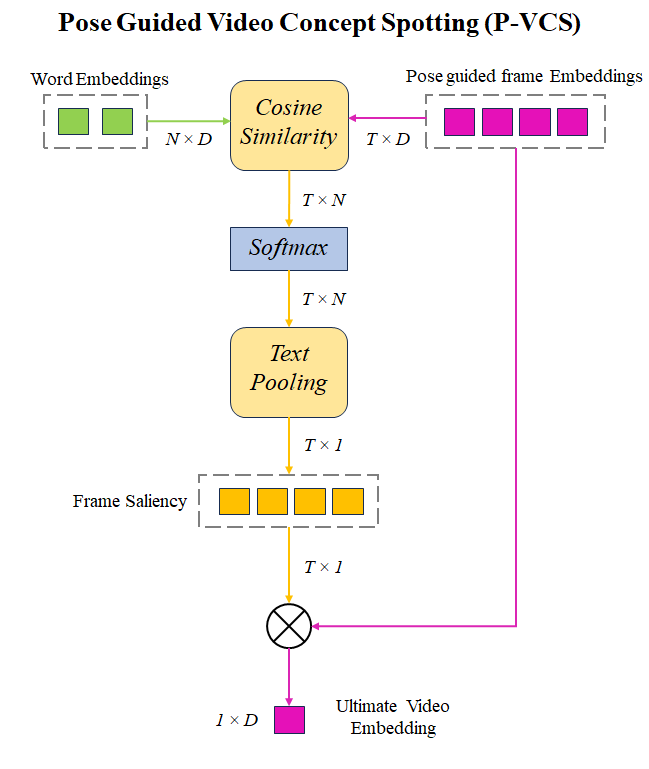}
    \caption{P-VCS: The pose-guided frame embedding is \textit{combined} with the word embedding to generate final video embedding.}
    \label{fig:p-vcs}
\end{figure}
\section{Methodology} \label{sec:method}
In this section, we get acquainted with the terminology which will be used throughout the paper and also elaborate on our proposed framework in more detail.

\subsection{Formulation of Problem and Terminology}
Given a video input \textit{v} and a collection of categories $\textit{C}=
\{c_1,c_2,...,c_\textit{M}\}$, where \textit{M} is the number of classes, the aim of action recognition task is to classify the video \textit{v} into a category \textit{c} $\in$ \textit{C}. Following \cite{bike}'s sampling scheme, \textit{T} frames are sampled from the video input \textit{v} and encoded using a vision encoder (represented by $f(\cdot|\theta_v)$) to obtain video embedding $\textbf{e}_v$. A text encoder $g(\cdot|\phi_c)$ is employed to encode the class \textit{c}
into category embedding $\textbf{e}_c$. Mathematically,
\begin{align}
    \textbf{e}_v=f(v|\theta_v),
    \textbf{e}_c=g(c|\phi_c)
\end{align}

The goal during model training is to maximise the similarity score between
$\textbf{e}_v$ and $\textbf{e}_c$ and minimise it in all other cases. The similarity score is denoted by:
\begin{equation}
    S_V=CS(\textbf{e}_v,\textbf{e}_c),
\end{equation}
where $CS(\cdot,\cdot)$ is the cosine similarity function.
During inference, the $S_V$ between the video embedding and each category embedding is calculated and the category with the highest score is selected as the top-1 prediction. We follow the architecture of BIKE's video branch for video and text encoding and utilize CLIP pre-trained ViT for video representation and CLIP's text encoder for textual context modeling respectively. Similarly, for our novel pose branch, we leverage CLIP pre-trained ViT for harnessing pose information. Throughout the rest of this work, we use these notations consistently.

\subsection{Pose Guided Video Concept Spotting (P-VCS)}
\textbf{Overview}. Typically, to obtain a video representation based on a pre-trained image model, the typical pipeline involves two stages: 1) Extracting the spatial embedding of each frame and, 2) Using an aggregation scheme to group these spatial features in order to incorporate temporal information inside them. We use the same pre-trained image model to extract spatial features from the sampled video frames and their corresponding pose heatmaps, which are tensors having a dimension of (H,W,1). The pose heatmap highlights all the key points in their respective frames and ultimately leads to better pose-augmented frame embedding, which is successively aggregated. Unlike \cite{bike}, where authors conducted an exploration to leverage the Text-to-Video knowledge through their Video Concept Spotting (VCS) mechanism, we propose the use of the 2D skeleton or pose-dependent temporal saliency conditioned on textual encoding to steer the temporal aggregation process and generate highly enhanced and relevant video representations.

\textbf{Pose heatmap generation.}
We follow openpose algorithm \cite{cao2017realtime} to obtain robust pose heatmaps of the corresponding frames. Specifically, we generate heatmaps with dimensions (H, W, 19), where each of the 18 channels represent one keypoint of the human body, and the 19th channel corresponds to the background. To convert this 19-channel heatmap into a tensor of dimension (H, W, 1), we extract the highest pixel value from the 18 corresponding pixels of the channels for each position and fill the (H, W, 1) heatmap accordingly. Additionally, the heatmap values are rescaled to ensure that the pixels lie within the range of 0 to 255.

\textbf{Pose augmentation in VCS}. In a formal context, the pre-trained Vision-Language Model (VLM) is designed to independently encode each element of the input, which includes videos, pose heatmaps, and category names. Subsequently, the model produces three distinct sets of embedding for these inputs: $\{ v_n \in \mathbb{R}^d \ | \ n = 1, 2, \ldots, T \}$ is the set of frame embeddings, $\{ p_n \in \mathbb{R}^d \ | \ n = 1, 2, \ldots, T \}$ is the set of pose embeddings, and $\{ t_k \in \mathbb{R}^d \ | \ k = 1, 2, \ldots, N \}$ is the set of word embeddings, where \textit{T}, \textit{N} and d are the number of sampled frames, the number of words in the class name and the encoding dimension respectively. We influence the frame embeddings by incorporating the pose embeddings using a straightforward attention mechanism involving the use of a sigmoid activation on the pose embeddings followed by an element-wise multiplication with it. This enables a seamless integration of pose information into the frame representation, enhancing the overall performance of the model. This is denoted by:
\begin{equation}
f_n=\sigma(p_n) \circ v_n,
\end{equation}
where $\sigma$ is the sigmoid function and $\circ$ denotes the hadamard product between $p_n$ and $v_n$. $f_n$ is the pose augmented frame embedding which captures much more prominent spatial information than the vanilla frame embedding.

To access the fine-grained relevancy, we compute the similarity between each word and every frame. Subsequently, a softmax operation is applied to normalize these similarities for each frame and by aggregating the normalized similarities of a specific frame with different words, we derive a frame-level importance score.
\begin{equation}
    S_n=\frac{1}{N} \sum_{k=1}^{N} \frac{\exp{(v_{n}^{\textbf{T}}t_k/\tau)}}{\sum_{n=1}^{T} \exp{(v_{n}^{\textbf{T}}t_k/\tau)}}, n \in [1,T], k \in [1,N],
\end{equation}
where $\tau$ is the temperature of this softmax function.
We utilise these temporal saliency scores to aggregate the pose guided frame embeddings as follows:
\begin{equation}
    \textbf{e}_v=\sum_{n=1}^{T} v_n S_n,
\end{equation}
where $\textbf{e}_v \in \mathbb{R}^d$ is the ultimate video representation. 
\begin{table*}[]
\caption{The comparison results with the existing SOTA methods on UCF-101 and HMDB-51 datasets. Here, \textcolor{magenta}{magenta} and bold font indicates the best performance without kinetics pre-training using ViT-B/16 backbone, whereas \textcolor{magenta}{magenta} and underline font indicates the best performance without kinetics pre-training using ViT-B/32 backbone. Similarly, \textcolor{blue}{blue} and bold (\textcolor{blue}{blue} and underline) font indicates the best performance using ViT-B/16 (ViT-B/32) backbone with pre-training.}
\centering
\begin{adjustbox}{width=0.9\textwidth}
\resizebox{\textwidth}{!}{%
\begin{tabular}{|c|c|c|c|c|}
\hline
\textbf{Training Mode} & \textbf{Method} & \textbf{Backbone} & \textbf{UCF-101} & \textbf{HMDB-51} \\ \hline
\multirow{8}{*}{\begin{tabular}[c]{@{}c@{}}Without \\ Pre-training\end{tabular}} & Two-Stream \cite{simonyan2014two} & - & 88.0 & 59.4 \\ \cline{2-5} 
 & IDT \cite{Wang_2013_ICCV}& - & 86.4 & 61.7 \\ \cline{2-5} 
 & Dynamic Image Networks + IDT \cite{Bilen_2016_CVPR} & - & 89.1 & 65.2 \\ \cline{2-5} 
 & TDD + IDT \cite{7299059} & - & 91.5 & 65.9 \\ \cline{2-5} 
 & BIKE \cite{bike} & \multirow{2}{*}{ViT-B/16} & 92.41 & 72.41 \\ \cline{2-2} \cline{4-5} 
 & ViLP (Proposed) &  & \textbf{\textcolor{magenta}{92.81}} & \textcolor{magenta}{\textbf{73.02}} \\ \cline{2-5}
 & BIKE \cite{bike} & \multirow{2}{*}{ViT-B/32} & 91.62 & 70.64 \\ \cline{2-2} \cline{4-5} 
 & ViLP (Proposed) &  & \textcolor{magenta}{\underline{91.78}} & \textcolor{magenta}{\underline{70.84}} \\ \hline
\multirow{6}{*}{\begin{tabular}[c]{@{}c@{}}With\\  Pre-training\end{tabular}} & Text4Vis \cite{text4vis}& \multirow{3}{*}{ViT-B/16} & 95.05 & 72.47 \\ \cline{2-2} \cline{4-5} 
 & BIKE \cite{bike} &  & 94.44 & 75.34 \\ \cline{2-2} \cline{4-5} 
 & ViLP (Proposed)&  & \textcolor{blue}{\textbf{96.11}} & \textcolor{blue}{\textbf{75.75}} \\ \cline{2-5} 
 & Text4Vis \cite{text4vis} & \multirow{3}{*}{ViT-B/32} & \textcolor{blue}{\underline{93.15}} & 70.84 \\ \cline{2-2} \cline{4-5} 
 & BIKE \cite{bike} &  & 91.22 & 70.16 \\ \cline{2-2} \cline{4-5} 
 & ViLP (Proposed)&  & 91.70 & \textcolor{blue}{\underline{70.98}} \\ \hline
\end{tabular}%
}
\end{adjustbox}
\end{table*}
We do action recognition by calculating the similarity score between the video embedding and the category embedding. This is done by performing the dot product between $\textbf{e}_v$ and $\textbf{e}_c$ as denoted by equation 2.
\subsection{Loss function}
Formally, our ViLP extracts feature representations $\textbf{e}_v$ and $\textbf{e}_c$, provided a video input \textit{v} and category \textit{c} with the corresponding encoders $f(\cdot|\theta_v)$ and $g(\cdot|\phi_c)$. Model parameters $\theta_v$ and $\phi_c$ are initialized with the weights from the pre-trained VLM (here, CLIP \cite{radford2021learning}).
While training, our aim is to guarantee that $\textbf{e}_v$ and $\textbf{e}_c$ are similar when they are related and dissimilar when they are unrelated.

Given a batch of \textit{B} triples $\{\textbf{e}_{vi}, \textbf{e}_{ci} \equiv C[y_i], y_i \}$, i ranging from 1 to \textit{B}, where \textit{C} is the collection of \textit{M} categories indexed by $y_i \in [0,\textit{L}-1]$, $y_i$ is a label indicating the index of the category, and $\textbf{e}_{vi}, \textbf{e}_{ci}$ denote the \textit{i}-th enhanced video embedding and category embedding respectively. We adopt a bidirectional learning objective and utilize a symmetric cross-entropy loss \cite{ju2022prompting,wang2021actionclip} to enhance the similarity between video-category pairs that match and to minimize the score for unrelated pairs:
\begin{gather}
    L_{V2C}=-\frac{1}{B} \sum_{i}^{B}\frac{1}{\lvert M(i) \rvert} \sum_{m \in M(i)} \log{\frac{\exp{(CS(\textbf{e}_{ci},\textbf{e}_{vm}/\tau)}}{\sum_{j}^{B} \exp{(CS(\textbf{e}_{ci},\textbf{e}_{vj}/\tau)} }},\\
    L_{C2V}=-\frac{1}{B} \sum_{i}^{B}\frac{1}{\lvert M(i) \rvert} \sum_{m \in M(i)} \log{\frac{\exp{(CS(\textbf{e}_{cm},\textbf{e}_{vi}/\tau)}}{\sum_{j}^{B} \exp{(CS(\textbf{e}_{cj},\textbf{e}_{vi}/\tau)} }},\\
    L_V=\frac{1}{2}(L_{V2C} + L_{C2V}),
\end{gather}
where $m \in M(i)=\{m \lvert m \in [1,B], y_m=y_i\}$, $CS(\cdot;\cdot)$ is the cosine similarity and $\tau$ refers to the temperature hyperparameter for scaling.
\section{Experimental Results}
\label{sec:results}
\subsection{Video Datasets}
\textbf{UCF-101} \cite{soomro2012ucf101} contains a total of 13,320 videos, distributed across 101 action categories. Each category represents a distinct human action or activity. The videos are sourced from various online video platforms, including YouTube, and were collected from a wide range of sources, making the dataset diverse and challenging.

The \textbf{HMDB-51} \cite{Kuehne2011HMDBAL} dataset consists of 7,000 video clips sourced from diverse origins, such as movies and web videos, resulting in a collection of realistic videos. The dataset is categorized into 51 action classes.

\textbf{Kinetics-400} \cite{kay2017kinetics} is an extensive video dataset, encompassing 240,000 training videos and 20,000 validation videos, covering a wide spectrum of 400 distinct human action categories. Each video clip in the dataset represents a 10-second segment of an action moment, expertly annotated from original YouTube videos.

We utilize UCF-101 and HMDB-51 datasets for both training and testing. However, we exclusively employ the Kinetics-400 dataset for pretraining purposes.

\subsection{Training Details}
In our experiments, we adopt the visual encoder of CLIP for video and pose encoding, and use the textual encoder for category representation. To prepare the video input, we sparsely sample T (=8) frames \cite{bike}. The temperature hyperparameter $\tau$ is set to 0.01 for all training phases. By default, the crop size is set to 224. We use \textit{AdamW} optimizer with a learning rate of 5e-5 (default for base models) and train base models for 30 epochs. Further, we resort to random flipping, random gray scale and random cropping augmentations for improving performance.

\subsection{Comparison Results} To better understand the performance of the proposed algorithm, we compare it with some of the existing SOTA methods. We found out that the proposed technique performs reasonably well even without the expensive Kinetics pretraining.  
In Table 1, we showcase our findings on the UCF-101 and HMDB-51 datasets and conduct a comprehensive comparison with state-of-the-art (SOTA) methods. Our evaluation encompasses two distinct training settings: one without video data pre-training and the other with pre-training, specifically using Kinetics as the video dataset. Our approach outperforms regular video recognition \cite{simonyan2014two,Bilen_2016_CVPR, Wang_2013_ICCV} by a substantial margin by employing only base Vision Transformers (ViTs) as backbone. We also demonstrate clear superiority over SOTA methods \cite{text4vis, bike} attempting to use Vision Language Pre-training (ViLP). Even without resorting to kinetics pre-training our model achieves 92.81\% Top-1 accuracy on UCF-101 and 73.02\% on HMDB-51 using ViT-B/16 as the backbone. With the utilization of kinetics pre-training and employing of ViT-B/16 backbone, our model achieves an impressive 96.11\% Top-1 accuracy on the UCF-101 dataset, surpassing the second-best \cite{text4vis} by more than 1.0\%. Our model, in this setting (ViT-B/16; kinetics pre-training) also outperforms all others on the HMDB-51 dataset, securing a commendable performance of 75.75\%. As mentioned in our key contributions, it is interesting to observe that with ViT-B/32 backbone, the proposed method performs better or closely on both datasets without pre-training compared to its pre-trained version. 

\subsection{Ablation Studies}
In this section, we provide extensive ablations to demonstrate the effectiveness of our pose augmentation. The outcomes of our investigation are presented in Table 2. It is important to note that to study the importance of a particular embedding, we keep the network architecture same, and train and infer the model by replacing the encoding under observation with a same-sized vector of constant value, irrespective of the input. 

\textbf{The Effect of Pose attention.}
Without the incorporation of the pose branch, the remaining modalities-video and text, the proposed architecture achieves 94.44\% and 75.34\% top-1 accuracies on UCF-101 and HMDB-51, respectively. Our final model, when equipped with Video, Language and Pose branches, has proven to outperform the SOTA methods, irrespective of the dataset and model backbone. For UCF-101, pose augmentation led to a gain of more than 1.5\% (ViT-B/16 setting; with pre-training) and more than 0.8\% for HMDB-51 (ViT-B/32 backbone; with pre-training) compared to the model trained only with Video and Text modalities. These results demonstrate the effectiveness of incorporating the Pose branch in our model for multi-modal tasks, surpassing the performance of the BIKE model across different scenarios.

\textbf{The Impact of Text branch.}
Notably, even in the absence of the text modality, our pose-guided video framework achieves comparable or superior results compared to the SOTA model reported in \cite{bike}. This highlights the effectiveness of using pose as a complementary modality for videos instead of language. By leveraging only video and pose information, our framework achieves an impressive accuracy of 95.56\% on the UCF-101 dataset and 75.35\% on HMDB-51, showcasing a substantial performance gain over their video and text augmented counterparts on these datasets as shown in Table 2. This further emphasizes the significant impact of incorporating pose information in enhancing the performance of multi-modal video action recognition tasks.

\textbf{The Impact of Video branch} We evaluated the performance of our model when using only the pose and text modalities, excluding RGB information. Surprisingly, the results were subpar, with an accuracy of 51.38\% on the UCF-101 dataset and 31.61\% on HMDB-51 dataset while employing the ViT-B/16 backbone with pre-training. This finding suggested that the pose and text modalities alone were not sufficient to achieve competitive performance on these datasets. This effect might be because of the fact that the video embedding captures additional information like the background and the scene, which are not captured in the text or embeddings but could be vital for decision-making. 

Our findings underscore the necessity of synergistically combining video, pose, and text modalities for video action recognition task. The results demonstrate that the integration of all three modalities is crucial to achieve optimal performance and robustness in this domain.
\begin{table}[]
\caption{Result of ablation study of modalities. All models consistently used ViT-B/16 backbone and were pre-trained with kinetics.}
\begin{tabular}{lcc}
\hline
\multicolumn{1}{c}{\textbf{Modes}} & \textbf{UCF-101} & \textbf{HMDB-51} \\ \hline
Pose + Text                        & 51.38            & 31.61            \\
Video + Text                & 94.44            & 75.34            \\
Video + Pose                       & 95.56            & 75.35            \\
Video + Pose + Text                & 96.11            & 75.75            \\ \hline
\end{tabular}
\end{table}
\section{Conclusion}
\label{sec:conclusion}
Through this work, we introduce a novel combination of pose, visual and text attributes that capitalizes on multi-modal knowledge to enhance video action recognition. Our approach involves the use of a pose encoder which captures valuable skeletal information in human action videos besides the typical vision and language models that are used to generate video and text encodings. Our pose augmented temporal saliency generation scheme, Pose Guided Video Concept Spotting (P-VCS) ensures that a superior video encoding is fabricated which eventually demonstrates its potency on two popular VAR datasets. The fusion of these three modalities convey new possibilities and potentials for enhancing video action recognition performance in thee future.

\bibliographystyle{ACM-Reference-Format}
\bibliography{citations}


\end{document}